%% file: main.tex
\documentclass[10pt,twocolumn,letterpaper]{article}

\usepackage{cvpr}              

\input{preamble}

\usepackage{multirow}

\definecolor{cvprblue}{rgb}{0.21,0.49,0.74}
\usepackage[pagebackref,breaklinks,colorlinks,citecolor=cvprblue]{hyperref}
\usepackage{pifont}
\newcommand{\xmark}{\ding{55}}
\usepackage{algorithm}
\usepackage{algpseudocode}  

\DeclareMathOperator*{\argmin}{arg\,min}
\usepackage{comment}

\def\paperID{22} 
\def\confName{CVPR}
\def\confYear{2024}

\title{Step Out and Seek Around: On Warm-Start Training with Incremental Data}

\author{Maying Shen, Hongxu Yin, Pavlo Molchanov, Lei Mao, Jose M. Alvarez\\
NVIDIA\\
{\tt\small \{mshen, dannyy, pmolchanov, lmao, josea\}@nvidia.com}}

\begin{document}
\maketitle
\input{sec/0_abstract}    
\input{sec/1_intro_new}

\input{sec/2_related_work}
\input{sec/3_method}
\input{sec/4_experiment}
\input{sec/5_summary}
{
    \small
    \bibliographystyle{ieeenat_fullname}
    \bibliography{main}
}


\end{document}

%% file: preamble.tex
\usepackage[dvipsnames]{xcolor}


%% file: sec/0_abstract.tex
\begin{abstract}
Data often arrives in sequence over time in real-world deep learning applications such as autonomous driving. When new training data is available, training the model from scratch undermines the benefit of leveraging the learned knowledge, leading to significant training costs.
Warm-starting from a previously trained checkpoint is the most intuitive way to retain knowledge and advance learning. However, existing literature suggests that this warm-starting degrades generalization. In this paper, we advocate for warm-starting but stepping out of the previous converging point, thus allowing a better adaptation to new data without compromising previous knowledge. We propose Knowledge Consolidation and Acquisition (CKCA), a continuous model improvement algorithm with two novel components. First, a novel feature regularization (FeatReg) to retain and refine knowledge from existing checkpoints; Second, we propose adaptive knowledge distillation (AdaKD), a novel approach to forget mitigation and knowledge transfer. We tested our method on ImageNet using multiple splits of the training data. Our approach achieves up to $8.39\%$ higher top1 accuracy than the vanilla warm-starting and consistently outperforms the prior art with a large margin. 
\end{abstract}

%% file: sec/1_intro_new.tex
\section{Introduction}
\label{sec:intro}

\begin{figure}
    \centering
    \includegraphics[width=\linewidth]{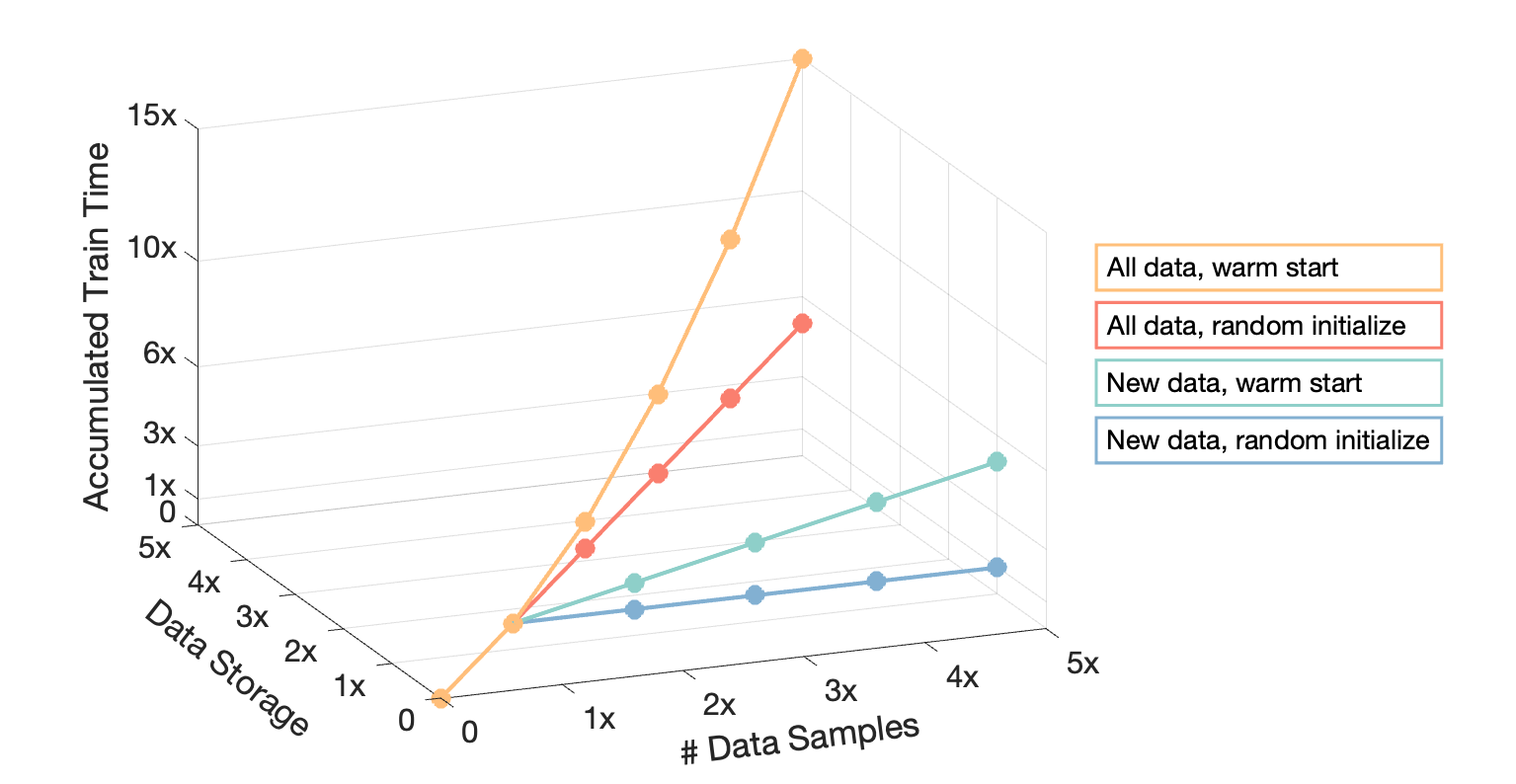}
    \caption{Accumulated training and storage costs as a function of the training strategy when training data is progressively available. Warm start involves using previously trained models and therefore increases the training costs. The goal here is to have fixed training and storage costs. That is, train a new model using existing models as initialization (warm-start) and only accessing to new data only.}
    \label{fig:train-cost}
\end{figure}

An open problem in deploying deep learning models in industry is enabling neural networks to learn incrementally from dynamic data streams. For instance, in autonomous driving, we want to improve the accuracy of a perception model systematically and efficiently as we obtain new training data~\cite{caccia2022anytime}. Continual learning, also called life-long or incremental learning, is the field of research focused on developing techniques and architectures that enable deep models to learn sequentially without needing to retrain them from scratch whenever new data is available. The academic setting usually focuses on incrementally learning different tasks~\cite{mallya2018packnet, sun2023decoupling}, continuously adapting to new domain data~\cite{garg2022multi}, or when we want to increment the number of categories/objects, the model can detect~\cite{luo2023class, sun2023decoupling}. In industry, however, the need is slightly different. We want a model to systematically leverage new training data for a fixed task and architecture to improve accuracy. We show how the accumulated training and storage costs change when data arrives sequentially under different training strategies in Fig.~\ref{fig:train-cost}. The straightforward approach to using the new data is to train the model from scratch with all available data. However, over time, this leads to prohibitive training costs. It is also impractical in storage-limited systems where storing all past data is not feasible. 
Thus, in this paper, we focus on developing efficient algorithms to update an existing model, boosting its accuracy using newly available data to achieve a constant training and storage cost.

The main challenge when improving, incrementally, a model's accuracy is finding the balance between forgetting and effectively retaining knowledge. On the one hand, we need to overcome catastrophic forgetting. That is, leverage the existing knowledge in the current model. On the other hand, as we might not have enough data, the model might have converged to a local minimum. Therefore, deciding what knowledge needs to be kept or discarded is crucial. 



Here, we introduce Knowledge Consolidation and Acquisition (CKCA), an algorithm for continuously learning from new data while consolidating knowledge from existing models. The algorithm has two main contributions: First, we propose feature regularization as means to refine new features using the previously learned ones and overcome catastrophic forgetting. In contrast to methods regularizing directly on the model's weights, our approach preserves relevant features while allowing the model's weights to better adapt to new data. 

Second, we introduce an adaptive knowledge distillation schedule. Our starting checkpoint is a weak teacher and can, therefore, be particularly relevant at the beginning of the training process. Instead of using a fixed distillation strength, we propose a decaying influence of the weak teacher on the student as the training progresses to prevent an over-coaching of the student.


Our experiments on ImageNet~\cite{imagenet} demonstrate that our approach yields up to $8.39\%$ higher accuracy compared to warm-starting the model and up to $6.24\%$ accuracy improvement compared to the current state of the art. Moreover, we can combine our method with existing approaches. For instance, combining our approach with iCaRL~\cite{rebuffi2017icarl}, a replay-based continual learning method, the accuracy is further boosted by $1.68\%$, and achieves up to $7.92\%$ higher than applying iCaRL only.

%% file: sec/2_related_work.tex
\section{Related Work}
\label{sec:related_work}

\textbf{Continual learning.} The need to adapt a trained model to exploit knowledge over a lifetime has motivated the study of continual learning. The major challenge is known as catastrophic forgetting, where adapting to new information results in reduced performance on old data. There are variants of continual learning according to the division of incremental batches~\cite{wang2023comprehensive} such as domain-incremental learning~\cite{garg2022multi}, task-incremental learning~\cite{mallya2018packnet, sun2023decoupling}, class-incremental learning~\cite{luo2023class, sun2023decoupling}, continual pre-training~\cite{sun2020ernie}, etc. 
Task-incremental and class-incremental learning~\cite{wang2023comprehensive} has received a lot of attention in the past years in a prosperous number of work. Among the current methods, the regularization-based approach~\cite{mirzadeh2020understanding, gao2023unified, kirkpatrick2017overcoming, kim2023achieving} is one of the most popular approaches. For example, EWC~\cite{kirkpatrick2017overcoming} constrains the model weights to stay close to their old values to maintain performance on old tasks while retaining the ability to learn new tasks; 
Replay-based methods are another thread of work that shows strength in continual learning. These approaches typically maintain a small memory buffer to store a few old training samples, which are carefully selected and updated to represent the old tasks such as ~\cite{rolnick2019experience, rebuffi2017icarl, yan2021dynamically}. Despite the tremendous number of work that has been proposed in this area, the problem of general data-incremental learning with network warm-starting is equally important in practice but understudied.

\textbf{Generalization on warm-starting.} Recent study on online learning and anytime learning~\cite{caccia2022anytime} has noticed the model's generalization degradation in the online setting. For example, the work~\cite{ash2020warm} revealed the finding that warm-starting from a pre-trained checkpoint, which is trained by partial data, leads to a lower accuracy than training the model from random initialization. To improve the model generalization, several weight reinitialization methods have been proposed~\cite{ash2020warm,zhou2022fortuitous,ramkumar2023learn}. The key to these methods is to forget partial of the learned knowledge selectively. For example, S\&P~\cite{ash2020warm} shrinks the magnitude of the learned weights and adds perturbation with small noises, preserving the relative activation but moving the network further from saturating regions. In LLF~\cite{zhou2022fortuitous} the authors propose to reinitialize the later layers of the model, memorizing the learned generalized low-level representation.
These methods mostly perform the experiments on small-scale datasets such as CIFAR datasets~\cite{cifar10}, with full access to previous data, but rarely explore the more practical and challenging cases with large-scale datasets and no replay of old data. 

%% file: sec/3_method.tex
\section{Method}
In this section, we introduce Continuous Knowledge Consolidation and Acquisition (CKCA), our approach to updating a model when new data becomes available without training the model from scratch. Our algorithm consists of three main components: Initialization, Regularization, and Distillation, see Algo.~\ref{algo:algorithm}. Bellow, we describe the algorithm for image classification, although it should be easily instantiable for segmentation of object detection.


\subsection{Preliminaries}
We denote a deep neural network model as $M$ encoded using the set of parameters $\Theta_0$. We assume a data distribution $D_{x, y}$ exists where $x$ is the input image and $y$ is the target output, e.g., label. For classification, $y\in\{1, ..., C\}$ where $C$ is the number of classes. We can also write the network output as $\hat{y}=M(\Theta_0, x)$.

We assume training data arrives in chunks in the form of a stream. For each stage $i$, we randomly sample $D_i$ following a distribution $D_{x, y}$. We define a data stream as an ordered sequence $S = \{D_1, ..., D_s\}$. Whenever we receive a new chunk of data $D_i$, we aim to train the model $M_i$ and ideally obtain $Acc_i > Acc_{i-1}$. We train $M_i$ at stage $i$ using $D_i^{train}=D_i$ whenever the previous training data is unavailable; Otherwise, the training data for stage $i$ is $D_i^{train}=\bigcup_{k=1}^i D_i$. We evaluate all models on a disjoint fixed test set $D^{test}$. That is, $D^{test}$ does not overlap with any training set.


\subsection{Initialization}
We use warm-start to initialize the weights of our model. That is, we use the weights of the previous model to initialize the model we want to train and, thus, inherit previously learned knowledge. Despite empirical evidence showing that warm-starting retains all previous knowledge and might hurt generalization~\cite{ash2020warm}, we will demonstrate in the experimental section that our approach mitigates this problem. 

\subsection{Feature Regularization - FeatReg}
The second module in our approach is a regularization term, \textbf{FeatReg}, added to the loss function during training. Existing approaches, such as elastic weight averaging~\cite{kirkpatrick2017overcoming}, regularize directly on the model's weights to prevent large deviations while enabling acquiring new knowledge either with minimum parameter variation or focusing on parameters identified as irrelevant. However, directly constraining the weights might limit the flexibility of the model. In addition, in our setting, the checkpoint has likely been trained with partial data, and therefore, identifying relevant parameters is misleading. In contrast, we constrain directly in the feature space, aiming to preserve relevant features while allowing the parameters to adapt to new data.

We formulate the loss function as, 
\begin{equation}
\label{eq:total_loss_full_access}
    L = \frac{1}{n}\sum_{(x,y)\in D_{i}^{train}} L_{CE}(\Theta, x, y) + \lambda L_{FeatReg},
\end{equation}
where $n$ is the size of $D_{i}^{train}$, $L_{CE}$ is the cross entropy loss, $L_{FeatReg}$ is the proposed feature regularization term, and $\lambda$ is a parameter to control the influence of the regularization term.

Given a checkpoint, we first identify key representative features $f_y$ for each class $y\in\{1, \dots, C\}$ and then, during training, penalize large deviations of the current features from representative ones. Specifically, we formulate $L_{FeatReg}$ as  
\begin{equation}
\label{eq:feat_reg}
      L_{FeatReg} \sim L_{reg} = ||g(\Theta, x) - f_y||_2, \ (x, y) \in D
\end{equation}
where $g(\cdot)$ extracts the feature representation of the input image $x$. In our experiments, $g(.)$ is the feature right before the classification layer. %

To obtain the key features for the $c$-th class, given the trained checkpoint $\Theta_{i-1}$ from previous stage, we first extract features for all samples belonging to class $c$ from the old dataset, then use $K$-means to cluster them into $K$ different groups $\mathcal{F}_{c, k}$, $1\leq k \leq K$, and, finally, compute the centroid for each group,
\begin{equation}
\label{eq:cls_feat_center}
    f_{c, k} = \frac{1}{n_{c,k}}\sum_{x\in \mathcal{F}_{c, k}} g(\Theta_{i-1}, x),
\end{equation}
where $n_{c,k}$ is the total number of samples in the group. In practice, we calculate and store the class feature centers when we train for model $M_{i-1}$.

When applying regularization during the new model $M_{i}$ training at the $i_{th}$ stage, we first load the stored class centroids. Then, for a data sample $(x, y)$ with feature map $g(\Theta, x)$, we constrain it to the closest class centroid,
\begin{equation}
    f_y = \argmin_{f_{c,k}|y=c} ||g(\Theta, x)- f_{c,k}||_2.
\end{equation}
In our experiments we use we use $K=1$ as we empirically found that there is significant different when using a larger number of centroids $K$.

\begin{algorithm}[t]
    \caption{Training of the model with CKCA}
    \label{algo:algorithm}
    \begin{algorithmic}[1]
        \State Initialize model $M$ by loading the previously trained model's weights $\Theta_{i-1}$. 
        \State Reset learning rate scheduler to regular train (as from scratch).
        \For{$iter = 1, \dots, max\_iter$}
            \State Calculate cross-entropy loss $L_{CE}$.
            \State Apply FeatReg and get $L_{FeatReg}$ as Eq.~\ref{eq:feat_reg}.
            \If{$D_i^{train}=D_i$} \Comment{with new data only}
                \State Apply AdaKD with adaptive strength as Eq.~\ref{eq:ada_strength}.
                \State Get total train loss as Eq.~\ref{eq:total_loss_full_access}.
            \ElsIf{$D_i^{train}=\bigcup_{k=1}^{i}D_k$} \Comment{with all data}
                \State Get total train loss as Eq.~\ref{eq:total_loss_no_access}.
            \EndIf
            \State Backpropagate and update model.
        \EndFor
        \State Calculate and store class feature center as Eq.~\ref{eq:cls_feat_center}.
    \end{algorithmic}
\end{algorithm}

\input{sec/main_result}

\subsection{Adaptive Knowledge Distillation - AdaKD}
\label{adakd}
Recall that our goal is to update a model using only new training data and a trained model using any previous training data, therefore, the main challenge is to prevent catastrophic forgetting\cite{lee2023study}. We use knowledge distillation between the previously trained model and the current one to transfer previously acquired knowledge during training, 
\begin{equation}
\label{eq:total_loss_no_access}
    L = \frac{1}{n} \sum_{(x,y)\in D_{i}} (1-\alpha)L_{CE} + \alpha L_{KD} + \lambda L_{FeatReg},
\end{equation}
\noindent 
where $L_{KD}=L_{KD}(M(\Theta, x), M(\Theta_{i-1}, x)| T)$ is the distillation loss, $T$ the temperature, $\alpha$ the distillation strength, and $M(\Theta_{i-1})$ is the model previously trained where we assume it has accumulated knowledge learned from the combination of $\{D_1, ..., D_{i-1}\}$. We set the distillation strength according to the data set size, 
\begin{equation}
    \label{eq:init_strength}
    \alpha_0 =\frac{\sum_{j=1}^{i-1}|D_j|}{\sum_{j=1}^{i}|D_j|}.
\end{equation}

Our goal is to systematically improve the performance of the model as new training data becomes available. Thus, it is licit to consider $M(\Theta_{i-1})$ is a weak teacher as it was trained with partial data. As such, we need to control its influence on the student as the training progresses. We propose to gradually decay the distillation strength $\alpha$ during training to make sure $M(\Theta_{i})$ is not ``over-coached'' by $M(\Theta_{i-1})$. Specifically, we set the initial distillation strength $\alpha_0$ as Eq.~\ref{eq:init_strength} and adaptively decrease the strength to $0.5$ at the end of training following a cosine decay function,
\begin{equation}
\label{eq:ada_strength}
    \alpha_t = \frac{2\alpha_0 - 1}{4}\cos{(\frac{t}{T}\times\pi)}+\frac{2\alpha_0 + 1}{4},
\end{equation}
where $t$ denotes the $t_{th}$ epoch and $T$ is the total number of training epochs, and we can get $\alpha_T=0.5$.

%% file: sec/main_result.tex
\begin{table*}[t]
    \centering
    \caption{Model performance on ImageNet $10$-splits \textbf{without} access to previous data.} 
    \label{tab:no_access_main_result}
    \resizebox{0.9\textwidth}{!}{
    \begin{tabular}{c|cccccccccc}
        \toprule
        \multirow{2}{*}{Method} & \multicolumn{10}{c}{Top1 Accuracy (\%)} \\
        \cmidrule{2-11}
         & Stage 1 & Stage 2 & Stage 3 & Stage 4 & Stage 5 & Stage 6 & Stage 7 & Stage 8 & Stage 9 & Stage 10 \\
         \midrule
        Train from scratch & $49.22$ & $49.44$ & $49.49$ & $49.58$ & $49.84$ & $49.57$ & $49.83$ & $48.53$ & $50.18$ & $50.01$ \\
        warm start~\cite{caccia2022anytime} & $49.22$ & $53.91$ & $56.42$ & $57.73$ & $58.30$ & $58.83$ & $59.09$ & $59.26$ & $59.80$ & $59.91$ \\
        S\&P~\cite{ash2020warm} & $49.22$ & $55.00$ & $57.30$ & $58.73$ & $58.85$ & $59.41$ & $60.11$ & $60.13$ & $60.59$ & $60.69$ \\
        LLF~\cite{zhou2022fortuitous} & $49.22$ & $52.16$ & $53.63$ & $54.55$ & $55.07$ & $55.46$ & $55.91$ & $56.09$ & $56.34$ & $56.40$ \\
        EWC~\cite{kirkpatrick2017overcoming} & $49.22$ & $54.04$ & $56.18$ & $57.64$ & $58.26$ & $58.91$ & $59.09$ & $59.28$ & $60.03$ & $60.17$\\
        iCaRL~\cite{rebuffi2017icarl} & $49.22$ & $55.36$ & $57.62$ & $59.35$ & $60.31$ & $60.70$ & $61.03$ & $61.42$ & $61.67$ & $62.06$\\
        Der++~\cite{yan2021dynamically} & $49.22$ & $54.88$ & $56.68$ & $58.01$ & $58.66$ & $59.21$ & $59.96$ & $60.21$ & $60.47$ & $60.56$ \\
        \textbf{$\text{CKCA}_{\text{FeatReg+AdaKD}}$ (Ours)} & $49.22$ & $\mathbf{57.92}$ & $\mathbf{61.08}$ & $\mathbf{62.87}$ & $\mathbf{64.24}$ & $\mathbf{65.42}$ & $\mathbf{66.40}$ & $\mathbf{67.05}$ & $\mathbf{67.57}$ & $\mathbf{68.30}$ \\
        \bottomrule
    \end{tabular}
    }
    \vspace{-0.5cm}
\end{table*}

%% file: sec/4_experiment.tex
\begin{table*}[t]
    \centering
    \caption{Our method can be combined with other approaches without conflict, yielding better accuracy. Experiments are performed on ImageNet $10$-splits with no access to previous data.} 
    \label{tab:combine_with_other_methods}
    \resizebox{\textwidth}{!}{
    \begin{tabular}{ccc|cccccccccc}
        \toprule
        FeatReg & \multirow{2}{*}{S\&P} & \multirow{2}{*}{iCaRL} & \multicolumn{10}{c}{Top1 Accuracy (\%)} \\
        \cmidrule{4-13}
        +AdaKD & & & Stage 1 & Stage 2 & Stage 3 & Stage 4 & Stage 5 & Stage 6 & Stage 7 & Stage 8 & Stage 9 & Stage 10 \\
         \midrule
        \checkmark & \xmark & \xmark & $49.22$ & $57.92$ & $61.08$ & $62.87$ & $64.24$ & $65.42$ & $66.40$ & $67.05$ & $67.57$ & $68.30$ \\
        \checkmark & \checkmark & \xmark & $49.22$ & $58.43$ & $61.68$ & $63.56$ & $64.53$ & $65.56$ & $66.42$ & $67.08$ & $67.56$ & $68.00$\\
        \checkmark & \xmark & \checkmark & $49.22$ & $58.33$ & $61.86$ & $64.08$ & $65.25$ & $66.39$ & $67.38$ & $68.25$ &  $68.66$ & $69.98$\\
        \checkmark & \checkmark & \checkmark & $49.22$ & $58.75$ & $62.38$ & $64.62$ & $65.92$ & $67.11$ & $67.76$ & $68.16$ & $68.68$ & $69.03$ \\
        \bottomrule
    \end{tabular}
    }
\end{table*}

\begin{table*}[t]
    \centering
    \caption{The model performance on ImageNet $10$-splits when previous data is available.} 
    \label{tab:full_access_main_result}
    \resizebox{0.9\textwidth}{!}{
    \begin{tabular}{c|cccccccccc}
        \toprule
        \multirow{2}{*}{Method} & \multicolumn{10}{c}{Top1 Accuracy (\%)} \\
        \cmidrule{2-11}
         & Stage 1 & Stage 2 & Stage 3 & Stage 4 & Stage 5 & Stage 6 & Stage 7 & Stage 8 & Stage 9 & Stage 10 \\
         \midrule
        Train from scratch & $49.22$ & $61.11$ & $66.71$ & $70.12$ & $71.94$ & $73.71$ & $75.01$ & $75.97$ & $76.71$ & $77.30$ \\
        warm start~\cite{caccia2022anytime} & $49.22$ & $59.96$ & $65.29$ & $69.22$ & $71.49$ & $73.43$ & $74.77$ & $76.20$ & $76.99$ & $77.61$ \\
        S\&P~\cite{ash2020warm} & $49.22$ & $61.64$ & $67.05$ & $70.58$ & $72.66$ & $74.18$ & $75.48$ & $76.42$ & $77.07$ & $77.83$ \\
        LLF~\cite{zhou2022fortuitous} & $49.22$ & $62.33$ & $67.67$ & $70.83$ & $72.95$ & $74.45$ & $75.40$ & $76.20$ & $77.04$ & $77.58$ \\
        EWC~\cite{kirkpatrick2017overcoming} & $49.22$ & $60.16$ & $65.53$ & $69.25$ & $71.75$ & $73.47$ & $74.81$ & $76.13$ & $76.97$ & $77.54$ \\
        \textbf{$\text{CKCA}_{\text{FeatReg}}$ (Ours)}  & $49.22$ & $\mathbf{63.01}$ & $\mathbf{68.00}$ & $\mathbf{71.45}$ & $\mathbf{73.41}$ & $\mathbf{74.84}$ & ${75.89}$ & ${76.65}$ & ${77.31}$ &${77.85}$ \\
        \textbf{$\text{CKCA}_{\text{FeatReg+AdaKD}}$ (Ours)}  & $49.22$ & $59.80$ & $65.76$ & $69.50$ & $72.45$ & $74.38$ & $\mathbf{76.02}$ & $\mathbf{77.02}$ & $\mathbf{77.49}$ & $\mathbf{78.09}$\\
        \bottomrule
    \end{tabular}
    }
\end{table*}

\section{Experiments}
In this section, we demonstrate the efficacy of the proposed algorithm using a ResNet50 on the ImageNet ILSVRC2021~\cite{imagenet} classification dataset. To mimic the data stream, we split the ImageNet training set randomly and evenly into multiple chunks. All the models are evaluated on ImageNet test set and the Top1 accuracy is reported. 

In all the experiments, we use the same recipe for training the models at any stage. While fine-tuning could lead to a faster adaptation, the learning rate needs to be tuned carefully at each stage for best results~\cite{ash2020warm, shon2022dlcft}. In contrast, using the same training recipe is straightforward and does not depend on the number of data streams available. The model and training setup applies to all the methods used for comparison.



\subsection{Results}
The goal of the first experiment is to evaluate our proposal when only new data and the previous checkpoint are available. That is, to train $M_i$ we only have access to $M_{i-1}$ and $D^{train}_{i}=D_i$ to obtain $Acc_i > Acc_{i-1}$.

For comparison, we consider six continual learning methods. First, as baselines, three warm start methods, the vanilla warm start, shrink and perturb (S\&P)~\cite{ash2020warm} and the recently published forget and relearn (LLC) method~\cite{zhou2022fortuitous}. Then, Der++~\cite{yan2021dynamically} and iCaRL~\cite{rebuffi2017icarl}, the two state-of-the-art replay-based methods proposed for task/class-incremental continual learning and we use memory size $20,000$ following the setup in the papers\cite{yan2021dynamically, rebuffi2017icarl}. We also include the well-known Elastic Weight Averaging (EWC)~\cite{kirkpatrick2017overcoming} as representative of regularization methods. For a fair comparison, we use the same initial checkpoint $M_0$ for all the methods. 

Tab.~\ref{tab:no_access_main_result} shows the results for this experiment on ImageNet with $10$-splits. The first thing we can observe is that S\&P outperforms vanilla warm-up. This is particularly relevant in low data regimes. However, as the training progresses, the gap between these two methods is reduced. Interestingly, forget and relearn~\cite{zhou2022fortuitous} does not improve generalization compared to vanilla warm-up. We can also observe that the accuracy of EWC~\cite{caccia2022anytime} is not significantly higher than vanilla warm-start. This is likely due to the fact that EWC assumes full model convergence at each stage which is not happening in our setting. Therefore, the selection of important parameters to be kept is likely biased towards local minima. In this setting, replay-based methods help improve the accuracy but the improvement is not significant. In contrast, our proposed method outperforms all the other methods in all the stages. In the early stages, we outperform all other methods with up to $4\%$ improvement compared to vanilla warm-start. More importantly, at the end of training, when the model has seen all the data, our approach yields an absolute improvement of $6.24\%$ compared to iCaRL~\cite{rebuffi2017icarl} and $8.39\%$ when compared to vanilla warm-start. 
Our approach uses regularization and distillation to accelerate the way models integrate newly acquired data. As such, our approach can be complementary to existing ones. In this experiment, we analyze the effect of combining our approach with two existing methods. First, we evaluate including shrink and perturb~\cite{ash2020warm} as a way to initialize the training. Second, the effect of also integrating iCaRL, the replay approach presented in~\cite{rebuffi2017icarl}. Tab.~\ref{tab:combine_with_other_methods} shows the results of this analysis. We can see two things. First, Shrink and Perturb by itself does not seem to bring any benefits compared to simply loading the previous checkpoint. However, there is a boost in performance when iCaRL is also added to the combination. In that case, we observe top-1 improvements across all training stages. Interestingly, the best results are obtained by combining our approach with iCaRL. In that case, we obtain more than $1\%$ top-1 accuracy improvement compared to using our method alone.


\begin{figure}[!t]
    \centering
    \vspace{-0.5cm}
    \begin{subfigure}{.5\linewidth}
    \centering
    \includegraphics[width=\linewidth]{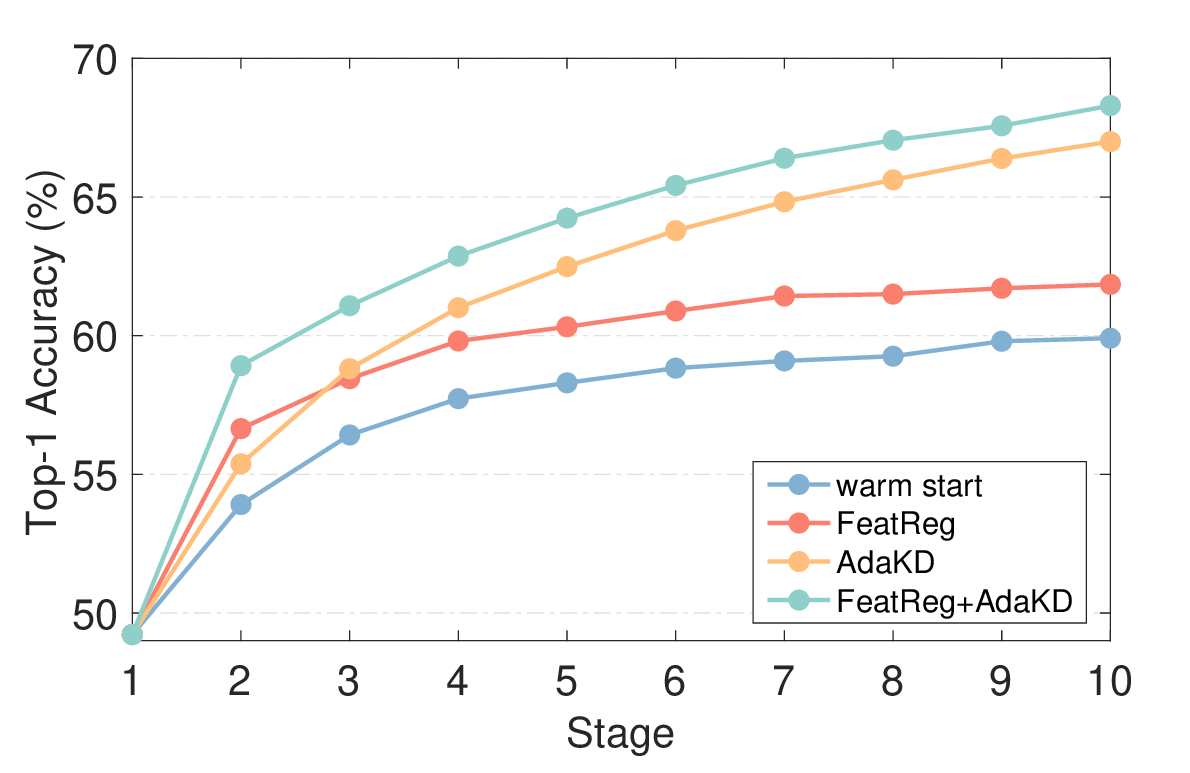}
    \caption{Old data is inaccessible}
    \end{subfigure}%
    \begin{subfigure}{.5\linewidth}
    \centering
    \includegraphics[width=\linewidth]{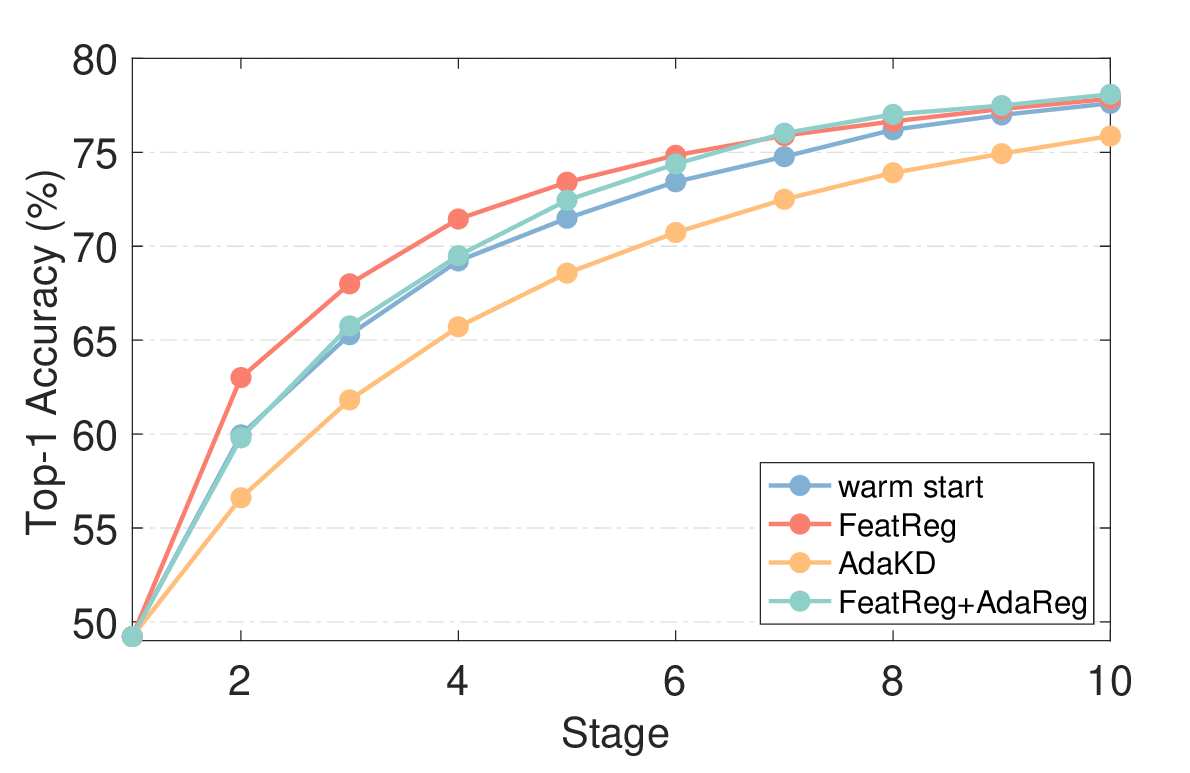}
    \caption{With full access to old data}
    \end{subfigure}%
    \caption{Accuracy improvement brought by FeatReg and AdaKD depending on the amount of data available.}
    \label{fig:featreg_adakd_ablation}
    \vspace{-0.35cm}
\end{figure}

In a second experiment, we relax the constraints and assume the data used to train previous checkpoints is still available. That is, to train $M_i$ we have access to $M_{i-1}$ and $D^{train}_i=\bigcup_{k=1}^i D_i$ and we aim to obtain $Acc_i > Acc_{i-1}$. In this case, as shown in~Fig.\ref{fig:train-cost}, the training costs increase significantly with the number of training stages. Tab.~\ref{tab:full_access_main_result} shows the summary of results for this experiment. In this case, we include two instances of our approach. One using feature regularization and adaptive knowledge distillation and a second instance where we disabled adaptive distillation. The first thing to observe, is that vanilla warm-start leads to significant accuracy degradation in the early stages of training. Interestingly, as the amount of available training data increases, warm-start recovers and performs slightly better compared to training the model from scratch. Early stage results are well aligned with other results reported in the literature~\cite{ash2020warm, ramkumar2023learn}. However, those existing works only report results on small datasets such as CIFAR-10, therefore there is no similar evidence for the late stages.
Our method, in contrast, outperforms all other methods towards the end of training. Interestingly, comparing the two instances of our method, we can conclude that the adaptive knowledge distillation module hurts performance in the early stages. Intuitively, adaptive knowledge distillation is less effective when the checkpoint acts as a weak teacher. In those cases, accessing directly to the data is more effective. As checkpoints improve performance, so does the effectiveness of our adaptive distillation module. 

\begin{figure}[!t]
    \centering
    \vspace{-0.5cm}
    \includegraphics[width=\linewidth, height=3.5cm]{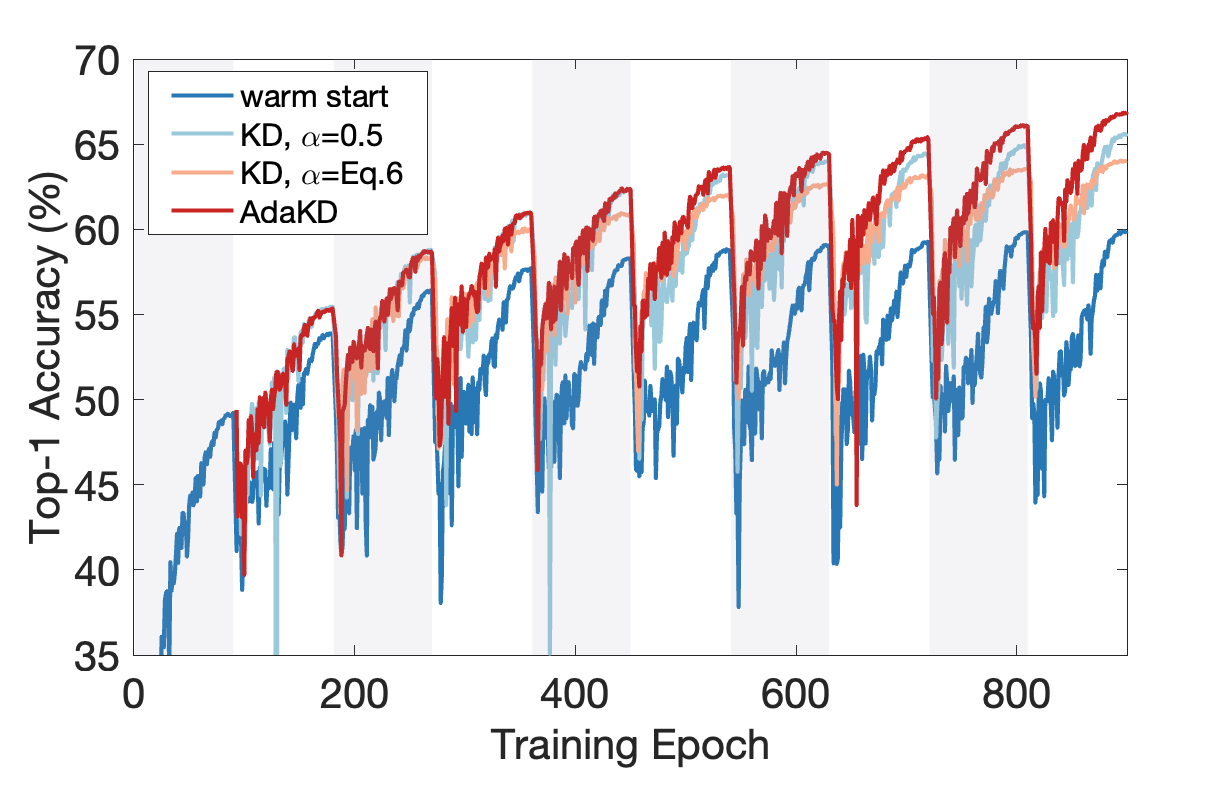}
    \caption{\textbf{Knowledge Distillation:} Comparison between the proposed adaptive distillation and other approaches. In early stages, all methods perform similarly. In later stages of training, however, our AdaKD clearly outperforms other approaches as we reduce the influence of the teacher when the student starts outperforming the teacher (checkpoint).}
    \label{fig:kd_ablation}
\end{figure}

\subsection{Ablation Studies}
We now analyze the contributions of the different modules of our approach to the final accuracy. To this end, we use ImageNet $10$-splits and report results when FeatReg and AdaKD are disabled for both cases: when we only have access to newly acquired data, and when previous data is available. As shown in Fig.~\ref{fig:featreg_adakd_ablation}, our feature regularization approach brings significant benefits in those cases where the amount of data is limited. Its effectiveness decreases as the amount of training data available increases. We also observe how adaptive regularization systematically mitigates forgetting by leveraging the existing checkpoints when old data is inaccessible. Those benefits are larger when we combine both models. The adaptive knowledge distillation module hurts the performance when all the data is accessible as we show in Tab.~\ref{tab:full_access_main_result}; but when combined with feature regularization, it still brings benefits to the later stage of training.


In the second ablation study, we analyze the behaviour of the adaptive component in knowledge distillation. For comparison, we consider vanilla distillation with two distillation strengths: $\alpha=0.5$ and $\alpha=\frac{\sum_{j=1}^{i-1}D_j}{\sum_{j=1}^{i}D_j}$. Fig.~\ref{fig:kd_ablation} summarizes the results for this analysis. As shown, the higher the strength, the faster the model learns during the early training stages. However, the gap decreases when the student performs on par or outperforms the teacher. When using a low distillation strength, the model learning is slower but ends up converging to a relatively higher accuracy. In contrast, our adaptive distillation approach combines the benefits of both maintaining a fast learning pace during the entire process and outperforming the other approaches especially in later stages when more data has already been processed. 



%% file: sec/5_summary.tex
\section{Conclusions}


We propose a novel approach to improve model performance and generalization with data available sequentially. We encourage the model to get out of the previous saturation point first and then seek a better one that captures refined feature representations by applying feature regularization. We also propose adaptive knowledge distillation to mitigate catastrophic forgetting. Experiments on ImageNet demonstrate the efficacy of our method. We achieve up to $8.39\%$ and $6.24\%$ accuracy improvement compared to the baseline and the best of prior arts, respectively.